\documentclass[conference]{IEEEtran}
\IEEEoverridecommandlockouts
\usepackage{cite}
\usepackage{amsmath,amssymb,amsfonts}
\usepackage{algorithmic}
\usepackage{graphicx}
\usepackage{textcomp}
\usepackage{xcolor}
\usepackage{subcaption} 
\def\BibTeX{{\rm B\kern-.05em{\sc i\kern-.025em b}\kern-.08em
    T\kern-.1667em\lower.7ex\hbox{E}\kern-.125emX}}
\begin{document}

\title{Point cloud obstacle detection with the map filtration
\thanks{The completion of this paper was made possible by the grant No. FEKT-S-20-6205 - "Research in Automation, Cybernetics and Artificial Intelligence within Industry 4.0" financially supported by the Internal science fund of Brno University of Technology.}}
\author{\IEEEauthorblockN{1\textsuperscript{st} Lukáš Kratochvíla}
\IEEEauthorblockA{\textit{Department of Control and Instrumentation} \\
\textit{Brno University of Technology}\\
Brno, Czechia \\
ORCID: 0000-0001-8425-323X}
}

\maketitle

\begin{abstract}
Obstacle detection is one of the basic tasks of a robot movement in an unknown environment. The use of a LiDAR (Light Detection And Ranging) sensor allows one to obtain a point cloud in the vicinity of the sensor. After processing this data, obstacles can be found and recorded on a map. For this task, I present a pipeline capable of detecting obstacles even on a computationally limited device. The pipeline was also tested on a real robot and qualitatively evaluated on a dataset, which was collected in Brno University of Technology lab. Time consumption was recorded and compared with 3D object detectors.
\end{abstract}


\section{Introduction}
Robotics is a sector that is experiencing a huge boom nowadays. Different types of industries e.g. automotive, aerospace, chemical, and others, as well as the social care field, are increasingly discovering, adopting, and seeking opportunities to robotize various human tasks. However, the main driver of progress is an industry. The need to automate manufacturing processes, optimize production, and move objects requires specialized equipment capable of working autonomously.

\begin{figure}[htbp]
    \centerline{\includegraphics[width=0.5\textwidth]{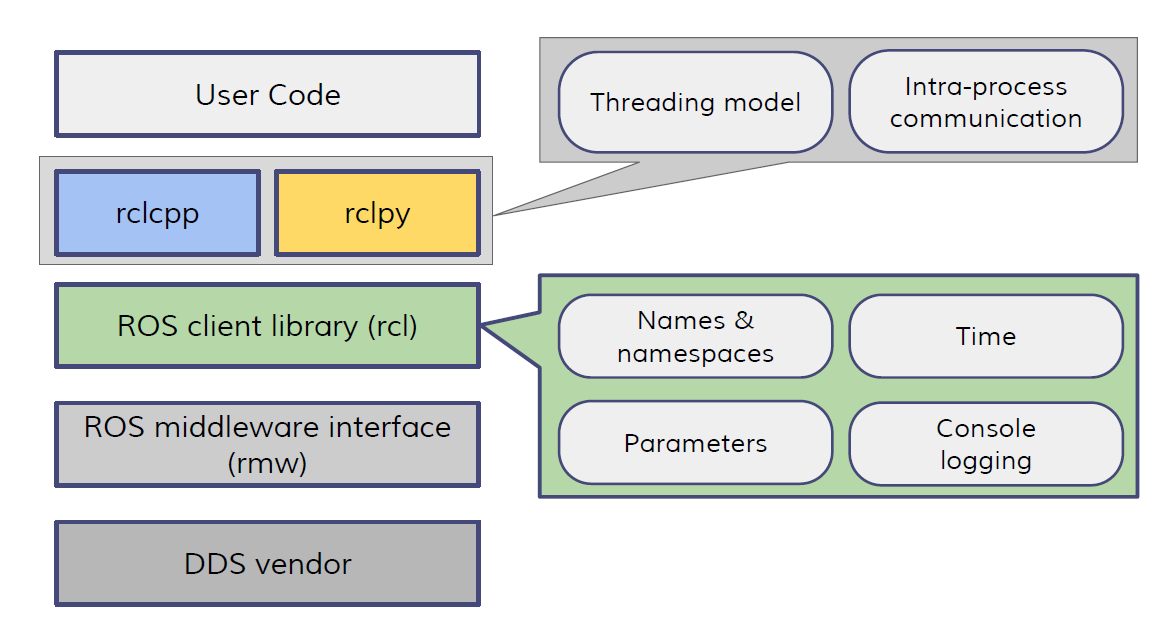}}
    \caption{The ROS architecture overview\cite{ros_a}. Whole library consists of five main parts. At the top of the pyramid is used code in C++ or python. Next level is responsible for translation of the user code. Level below is client library with name-spaces, logging, parameters and time clocking. On the bottom is communication provider.}
    \label{fig:ros}
\end{figure}

One of the first steps towards this goal is to create a robot capable of avoiding obstacles. The robot must not only sense its environment but also understand it, map it, be able to navigate through it, and react to changes in it. For these tasks, the ROS (Robot Operating System) environment has been created to implement different nodes that can communicate with each other and transfer data. ROS architecture is depicted on Fig~\ref{fig:ros}. An essential part of such a robot capable of avoiding obstacles is a localization and navigation system. If the robot could not locate itself, it would not be able to navigate through the map and perform the tasks given. On the other hand if the robot localizes itself and knows where it is going, it must be able to map new obstacles it encounters and avoid them if necessary.

There are several ways in which we can perceive the robot's environment. Some rely on visual information, in which a large amount of data is hidden, but it is necessary to understand what is in the scene. Others explore the surrounding environment directly using distance-measuring sensors such as LiDAR, stereometry or ultrasound. This paper focus on data from a LiDAR sensor.

The work is structured as follows. The section~\ref{sec:ra} describes papers dealing with similar topics. The section~\ref{sec:et} describes methods used in a new pipeline. The pipeline design in the following section~\ref{sec:pp} is proposed based on this research. The section~\ref{sec:e} discusses the implementation of the pipeline and the results in the experimental setup. The section~\ref{sec:c} summarizes the achievements of the whole paper.

\section{Related articles}
\label{sec:ra}
Various methods are used to detect objects in a point cloud and some will be mentioned. Currently, the development focus on deep learning models. The models can be divided into two categories, i.e. single-step and multi-step.

The first group \cite{voxelnet,second,pointPillars,pointGnn,cbgs} works directly with the raw point cloud and searches for various objects in it. \cite{voxelnet} builds detection on feature extraction, convolutional layers and finding suitable candidates for output. \cite{second} takes a similar approach but modifies convolution for sparse input data. \cite{pointPillars} leverages previous experience from 2D SSD (Single Shot Detector) model. \cite{pointGnn} uses graphical representation of data for computing output.

The second group \cite{Shaoshuai,pointRCNN,STD,PointNet} performs multiple steps to detect an object. \cite{Shaoshuai} first detects parts of the object and then aggregates these parts to outputs. \cite{pointRCNN} also performs two steps, first detecting based on the raw data and second fine-tuning the detections with a second model. \cite{STD} uses model consisting of three steps. In the first step, features are generated based on the defined structures. In the second step, information is compressed into a more compact representation, and in the last step, detections are sought. \cite{PointNet} based on structured features, it uses them to create segmentation and classification of objects.

There is also a group of methods that try to achieve detection using classification on a 2D image. \cite{Mausavian,Buyu,Jason} try to use classification on an image to achieve 3D detection. \cite{Mausavian} create a representation of the image using attributes from a convolutional network. Afterwards it calculates the angle of the object, the size of the object in 3D, and the confidence of detection based on these features. \cite{Buyu} use a fully connected network to estimate the 2D detection, orientation of the object (direction for expansion to 3D), make a cubic assumption of the 3D object and extract the features. In the second step, they extract the 3D detection using a 3D model based on the extracted features. \cite{Jason} try to reconstruct objects in 3D from 2D image and create 3D detections based on them. 

\section{Explanation of terms}
\label{sec:et}
The \textit{Voxel Grid} is a representation of points in space that allows for a significant way to enable point manipulation. This representation divides space into small blocks and each block is represented by a center point. If the number of points in a given block is sufficient, the block will be implemented in the Voxel Grid, otherwise it will not. This method also allows to perform data filtering. This method significantly reduce number of points in a point cloud. Therefore it is absolutely necessary if we want to perform the calculations on a power-limited device.

\textit{The RANSAC} (Random Sample Consensus) \cite{ransac} method is one of the simplest methods for finding correspondence points. The method selects a random subset of the data and tries to find a pattern, then tests the remaining points and calculates the error. This procedure is repeated multiple times and for the result is chosen a model with the minimal error. This method can also be used to find surfaces in a point cloud.

\textit{KDtree} (short for k-dimensional tree) \cite{kd_tree} is a method that uses binary trees to represent the data and makes it easier to work with the data. The method is based on creating a binary tree that ideally captures the data. The search in this representation is simpler than in an unsorted data.

\textit{FLANN} (Fast Library for Approximate Nearest Neighbors) \cite{flann} is used to find the clusters using KDtree. FLANN is a library implementing the nearest neighbor approximation algorithm in a multidimensional space. It also allows to find the optimal cluster with the help of a projection into the floor plane of the object.

\begin{figure}[htbp]
    \centerline{\includegraphics[width=0.5\textwidth]{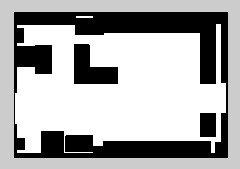}}
    \caption{A map example. This example of lab map was created with graphic editor.}
    \label{fig:map}
\end{figure}

\section{Proposed pipeline}
\label{sec:pp}
Many methods process point clouds, others follow the 3D object reconstruction route. As can be deducted from related work, all of these approaches require significant computational power. Therefore, in this section, I will propose a pipeline implementing obstacle detection given that it is desired to implement this task on a computationally constrained device.

The point cloud data will be first preprocessed. For preprocessing, methods for outlier removal, data reduction and floor removal will be used. Voxel Grid will be used for the first two tasks. This method is simple enough to use on the entire point cloud and brings a significant speedup in the entire post-processing. The RANSAC method will be used to find the floor. This method is slightly computationally intensive, so we could consider a fixed floor separation, but we could lose information about, for example, a hole in the surface. Subsequently, the data will be filtered based on the occupancy map to keep the new observations up-to-date.  The last point of the pipeline will be a cluster search that will group points into blocks. It will also allow to find out the ideal rotation of the blocks based on observations, so that they do not occupy space that is not observed.

The map can be created in advance, from knowledge of the space, or it can be created in progress from sensing the environment using the ROS package \textit{navigation2}. The Fig~\ref{fig:map} depicts a map of a sensed lab. For the map generation, the map created from the occupancy cells will be used. Occupancy grid is \textit{navigation2} package format for a map. The map must have a known transformation to the sensor and robot reference frame. The robot can modify the map as it moves by incorporating new obstacles.  

\begin{figure}[htbp]
    \centerline{\includegraphics[width=0.5\textwidth]{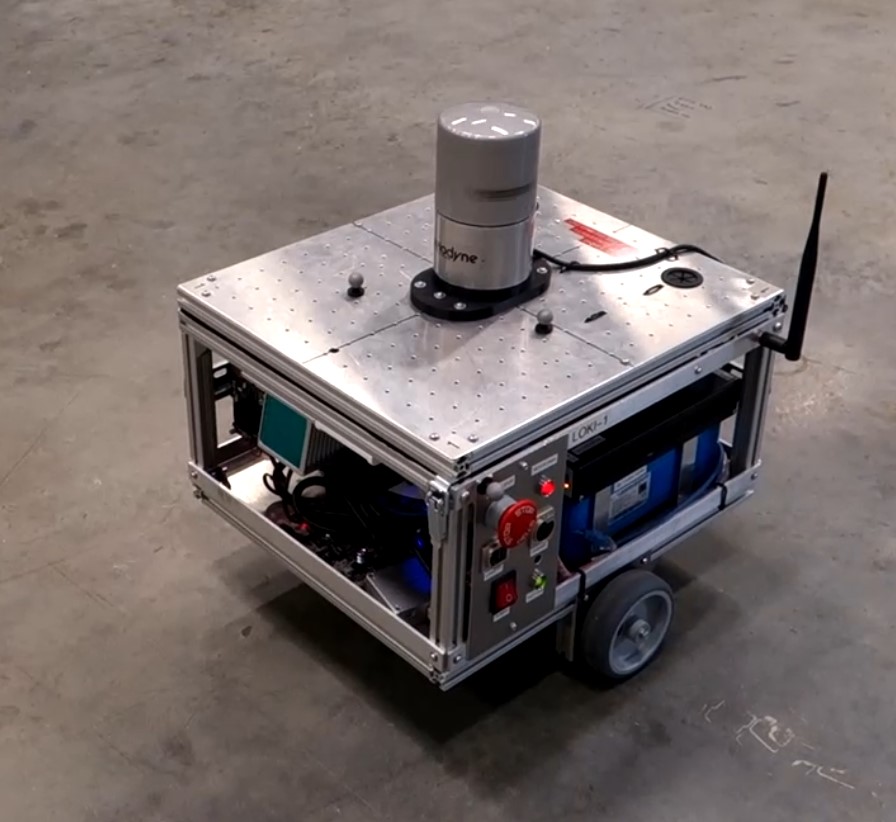}}
    \caption{The robot overview. LiDAR MID-70 is in front. Raspberry Pi is hidden inside. Battery and WiFi router are on the side. The robot was equipped with more facilities than is described in this paper.}
    \label{fig:robot}
\end{figure}

\section{Experiment}
\label{sec:e}
In this work, I have performed an experimental implementation of the introduced pipeline. For the experimental setup, I chose ROS2 foxy distribution\cite{ros2}, because it is currently supported version of ROS. It was installed on the Raspberry Ri with Ubuntu 18.04 OS (Operating System). The whole pipeline was attached to a robot that could be controlled remotely with a temporary power source. The goal was to achieve real-time detection of obstacles in the robot's surroundings as it moves, while recognizing objects located in front of the robot for timely stopping in case of a collision within the robot's trajectory. The robot is depicted on Fig~\ref{fig:robot}.

\subsection{Hardware}
Pipeline consisted of a MID-70 sensor from Livox and a Raspberry Pi computing unit with 4GB RAM. The MID-70 is LiDAR sensor, which consists of a laser beam source and a spreading mirror that produces a random scan of the area. The sensor senses a cone with a 70 degree FOV (Field Of View) and is capable of sensing from 5cm to tens of meters depending on the tilt of the installation and the reflectivity of the material being sensed. The Raspberry Pi is a credit card sized computing unit capable of running an OS with connectors for HDMI (High-Definition Multimedia Interface), USB (Universal Serial Bus), Ethernet, CSI (Camera Serial Interface), DSI (Display Serial Interface) and GPIO (General-Purpose Input/Output).

Furthermore, the robot was constructed as a two-wheeled machine with a differential chassis. A battery was used as the power source. A wifi-router was added on the robot for communication with the control. Robot was also equipped with more facilities, but they are not important for this work.

\begin{figure}[htbp]
    \centerline{\includegraphics[width=0.5\textwidth]{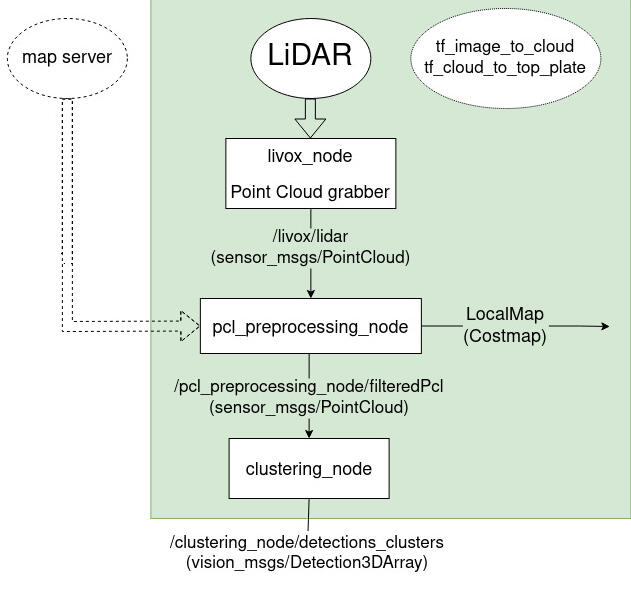}}
    \caption{Diagram with pipeline structure. The whole pipeline is computed on a computationally constrained device. In this diagram one can see messages used for communication between nodes.}
    \label{fig:struct}
\end{figure}

\subsection{Dependencies}
Following ROS2 packages were used. Package \textit{navigation2} for navigation and working with a map. Package \textit{tf2} for transformation from one frame to another and \textit{rviz2} for message vizualization. You can see the whole pipeline diagram in Fig~\ref{fig:struct}.

\subsection{Software}
Each node was created as a separate node in the ROS2 environment. This concept allows to implement differently demanding operations on different HW, or it allows to run a pipeline with multiple different settings. To implement the clustering and data preprocessing methods I have used the PCL (Point Cloud Library)\cite{pcl_library}. The PCL is open-source library specialized to work with point cloud.

\begin{figure*}[htbp]
    \centering
     \begin{subfigure}[c]{0.329\textwidth}
        \centering
         \includegraphics[width=\textwidth]{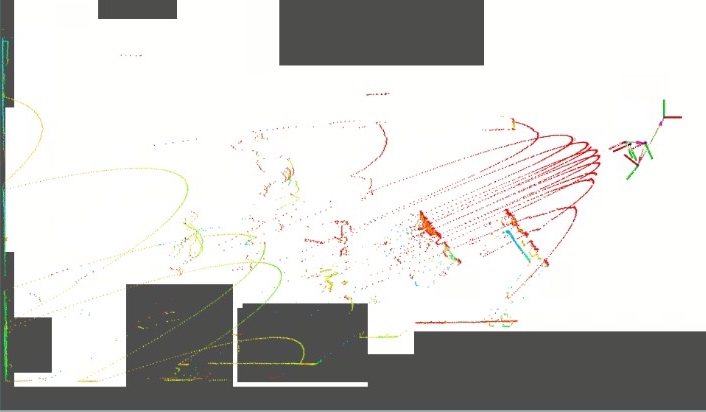}
         \caption{}
         \label{fig:raw_o}
     \end{subfigure}
     \hfill
     \begin{subfigure}[c]{0.329\textwidth}
         \centering
         \includegraphics[width=\textwidth]{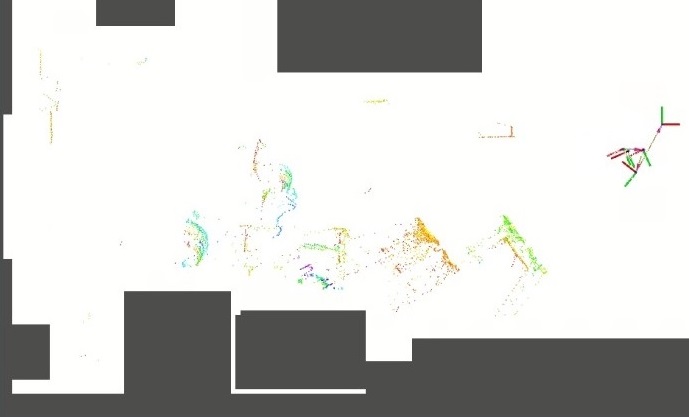}
         \caption{}
         \label{fig:pre_o}
     \end{subfigure}
    \hfill
     \begin{subfigure}[c]{0.329\textwidth}
         \centering
         \includegraphics[width=\textwidth]{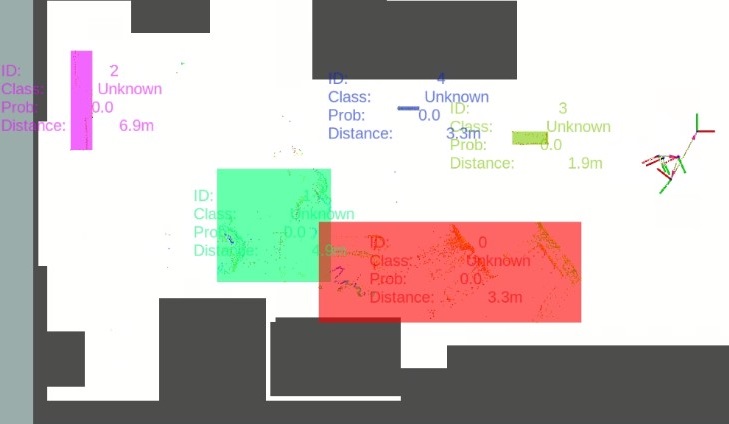}
         \caption{}
         \label{fig:det_o}
     \end{subfigure}
     \begin{subfigure}[c]{0.329\textwidth}
        \centering
         \includegraphics[width=\textwidth]{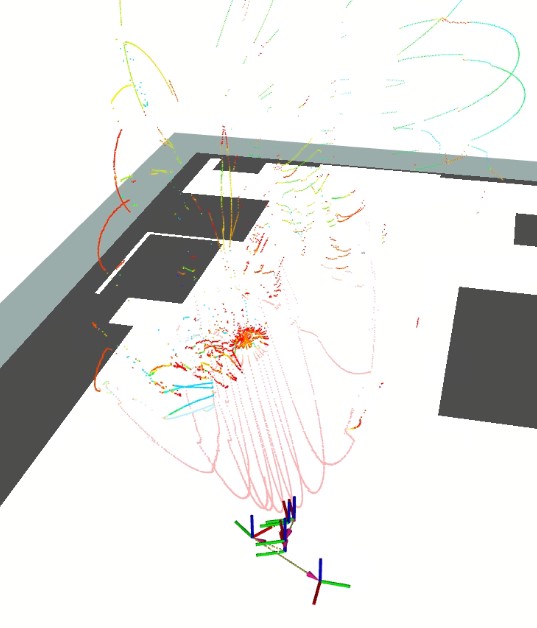}
         \caption{}
         \label{fig:raw}
     \end{subfigure}
     \hfill
     \begin{subfigure}[c]{0.329\textwidth}
         \centering
         \includegraphics[width=\textwidth]{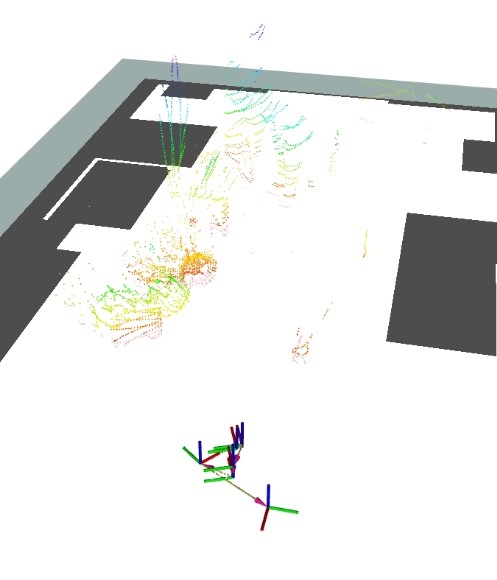}
         \caption{}
         \label{fig:pre}
     \end{subfigure}
    \hfill
     \begin{subfigure}[c]{0.329\textwidth}
         \centering
         \includegraphics[width=\textwidth]{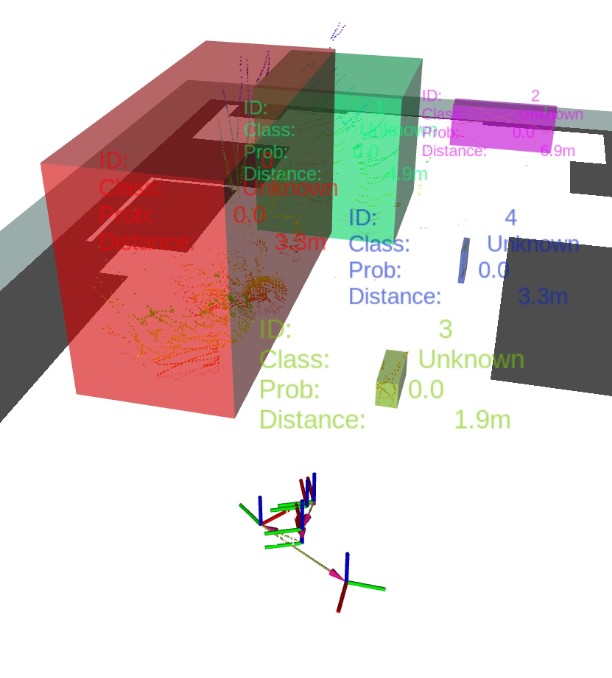}
         \caption{}
         \label{fig:det}
     \end{subfigure}
     \caption{Example of processed point cloud from two perspectives. \textit{(a)} Raw data from sensor (bird's-eye view), \textit{(b)} preprocessed data with \textit{pcl\_preprocessing\_node} (bird's-eye view), \textit{(c)} detections found by \textit{cloustering\_node} (bird's-eye view), \textit{(d)} Raw data from sensor (perspective view), \textit{(e)} preprocessed data with \textit{pcl\_preprocessing\_node} (perspective view), \textit{(f)} detections found by \textit{cloustering\_node} (perspective view).}
    \label{fig:com}
\end{figure*}

The \textit{livox\_ros2\_driver} was used as the first node that collects the sensor data. This software is created directly by Livox as a support for developers. It allows to collect data from multiple sensors at the same time. It puts the data into ROS messages that have a clear indication of which frame of reference they come from and what their timestamp is. The node calculates the timestamp from the timestamp differences of the incoming packets. The visualized raw data are depicted in Fig~\ref{fig:raw_o} and Fig~\ref{fig:raw}.

The next node in the sequence is \textit{pcl\_preprocessing\_node}. This node is responsible for appropriately processing the data so that clustering can be performed afterwards. The node can detect and remove the floor by finding the floor equation. Subsequently, it allows to reduce the amount of points in the point cloud by using a Voxel Grid. It also allows points to be transformed, with \textit{tf2} package, from one reference frame to another and filtered according to a map (in which known obstacles are entered). The point clouds are filtred with map by comparison of each cloud point with corresponding point in map. If the point in map is occupied, the point is removed from point cloud. The node also allows the creation of a local map, where the sensed part (of the environment only) is included. The input to this node is ROS messages from the previous node. The visualized preprocessed data are depicted in Fig~\ref{fig:pre_o} and Fig~\ref{fig:pre}.

The last node in the pipeline is \textit{clustering\_node}. This node allows to create objects from clusters in a point cloud by use of FLANN. The object is rotated around the center point and the content of the projected rectangle is correspondingly recomputed. When the minimum content is found, the remaining points are recalculated. The input to this node is preprocessed data and the output is ROS messages containing detections. The visualized detections are depicted in Fig~\ref{fig:det_o} and Fig~\ref{fig:det}.

I have added a visualization node to the entire pipeline for rendering using the \textit{rviz2} package. This node allows to convert non-visualizable (DetectionArray) messages into visualizable (MarkerArray) ones.
One node that stands aside from our pipeline but is necessary for pipeline implementation is \textit{map\_server}. It is responsible for working with the map. Receive changes from other nodes and provide the map on request. This node can create a map from a predefined one, or based on messages from the pipeline. If the obstacles disappear it is necessary to reboot this node.

An important aspect of the implementation was the creation of configuration files for easy parameter modification and the creation of startup scripts. Each node has its own parameters that are editable in a single configuration file. The startup scripts allow the configuration files to be loaded and it also allows individual nodes to be started.

\subsection{Results}
Several experiments were performed in the laboratory of Brno University of Technology. The aim of the experiment was to move the robot around the lab, detect obstacles and capture data for evaluation. A sensor was attached to the robot along with a computing unit, power source and controls. The following scenarios were prepared: sudden obstacle detection, static obstacle and map error. 

\begin{table}[htbp]
    \caption{Comparison of time consumption for two selected model and my pipeline. Note that both 3D object detectors were processed on GPU (Graphical Processing Unit), but pipeline was computed on CPU (Central Processing Unit)}
    \begin{center}
        \begin{tabular}{|c|c|}
            \hline
            \textbf{Method}&\textbf{Time[ms]} \\
            \hline
            CBGS\cite{cbgs} (on GPU) & 40 \\
            \hline
            Second\cite{second} (on GPU) & 171 \\
            \hline
            Pipeline (My on CPU)& 11 \\
            \hline
        \end{tabular}
        \label{tab1}
    \end{center}
\end{table}

The sensing frequency of the LiDAR was 10~Hz. A total of 17,000 point cloud samples were measured during 3 measurements. These samples were qualitatively evaluated, e. g. depicted in Fig~\ref{fig:com}. In Tab~\ref{tab1} is presented time consumption for two selected 3D object detectors and my pipeline. The pipeline allows to find obstacles in point cloud.

\section{Conclusion}
\label{sec:c}
The aim of this paper was to familiarize the reader with the possibilities of finding obstacles in space using a LiDAR sensor. These sensors are usually used in robotics to sense the surrounding environment. The methods that can be used are discussed. E. g. 3D object detectors could be use for obstacle searching, If we have enought computational power. On the other hand I want to search for obstacles on a computationally constrained device a new pipeline has been proposed. The pipeline consists of data preprocessing and clustering. The proposed pipeline was qualitatively verified on measured data. In the experiment section, the implementation of such a method was described along with the achieved results. The pipeline was tested on a computationally constrained device and also compared with 3D object detectors. The pipeline is suitable for obstacle search in real-time. 

Follow-up work will focus on extending the pipeline to include obstacle classification. The development will look at different responses to different types of obstacles. If a robot passes a moving human it should react differently than if it is moving around a shelf.


\end{document}